%% file: ApexFuse.tex
\documentclass[10pt,twocolumn,letterpaper]{article}

\usepackage[pagenumbers]{cvpr}      

\usepackage{booktabs}
\usepackage{colortbl}

\definecolor{bestcolor}{RGB}{255,180,180}
\definecolor{secondcolor}{RGB}{225,215,240}

\usepackage{algorithm}
\usepackage{algpseudocode}


\definecolor{cvprblue}{rgb}{0.21,0.49,0.74}
\usepackage[pagebackref,breaklinks,colorlinks,allcolors=cvprblue]{hyperref}

\def\confName{CVPR}
\def\confYear{2027}

\title{Beyond Resemblance: Image Fusion Reimagined}

\author{
Hao Zhang$^{1,2}$\thanks{Equal contribution} \quad
Yumin Huang$^{1*}$ \quad
Linfeng Tang$^{3}$ \quad
Junjun Jiang$^{4}$ \quad
Jiayi Ma$^{1,3}$\thanks{Corresponding author}\\
$^{1}$Electronic Information School, Wuhan University, Wuhan, China\\
$^{2}$Suzhou Institute of Wuhan University, Suzhou, China\\
$^{3}$School of Robotics, Wuhan University, Wuhan, China\\
$^{4}$School of Computer Science and Technology, Harbin Institute of Technology, Harbin, China\\
{\tt\small \{zhpersonalbox,linfeng0419,jyma2010\}@gmail.com, hym1024@whu.edu.cn, jiangjunjun@hit.edu.cn}
}

\begin{document}
\maketitle
\input{sec/0_abstract}

\input{sec/1_intro}

\input{sec/2_methodology}

\input{sec/3_experiments}

\input{sec/4_conclusion}
\input{sec/5_acknowledgments}

{
    \small
    \bibliographystyle{ieeenat_fullname}
    \bibliography{ref}
}

\end{document}

%% file: sec/0_abstract.tex
\begin{abstract}
Image fusion often lacks ground-truth fused images, forcing existing deep methods to indirectly define fusion objectives by requiring the fused image to simultaneously resemble multiple source images. However, the distinct imaging characteristics of different modalities can impose competing constraints on the fused result, potentially limiting the effective integration of their complementary information. To address this long-standing challenge facing the image fusion community, we propose ApexFuse, a novel generative image fusion paradigm that moves beyond source-resemblance constraints toward learning a general distribution of high-quality scene representations. Specifically, we develop a human-LMM-in-the-loop pipeline that leverages large multimodal generative models and iterative human feedback to translate abstract fusion objectives into reliable visual references, thereby defining the target distribution of ideal fusion outcomes. Through target-distribution-guided adaptation, we transfer the rich generative priors of pretrained FLUX to image fusion, enabling high-quality scene reconstruction conditioned on multimodal observations. Extensive experiments demonstrate that ApexFuse harmoniously integrates salient thermal targets, natural colors, and fine textures, achieving superior visual quality as well as semantic segmentation and object detection performance over previous source-resemblance methods. Moreover, ApexFuse exhibits remarkable cross-task generalization, extending from infrared-visible fusion to multi-exposure fusion, multi-focus fusion, and visible-depth fusion without additional fine-tuning.
\end{abstract}

%% file: sec/1_intro.tex
\vspace{-0.3cm}
\section{Introduction}\label{sec:intro}
\vspace{-0.1cm}
Image fusion aims to integrate complementary information from multiple imaging modalities into a unified representation, enabling more comprehensive and reliable scene perception for both human observers and intelligent machines~\cite{zhang2021image}. As a representative task, infrared-visible image fusion combines the thermal saliency of infrared imaging with the rich textures of visible imaging, enhancing perception under nighttime and low-visibility conditions for critical applications such as assisted driving, military reconnaissance, and intelligent surveillance~\cite{zhang2023visible}. However, despite substantial progress, a fundamental question remains unanswered: \textit{What should an ideal fused image look like}?

\begin{figure}[t]
	\centering
	\includegraphics[width=\columnwidth]{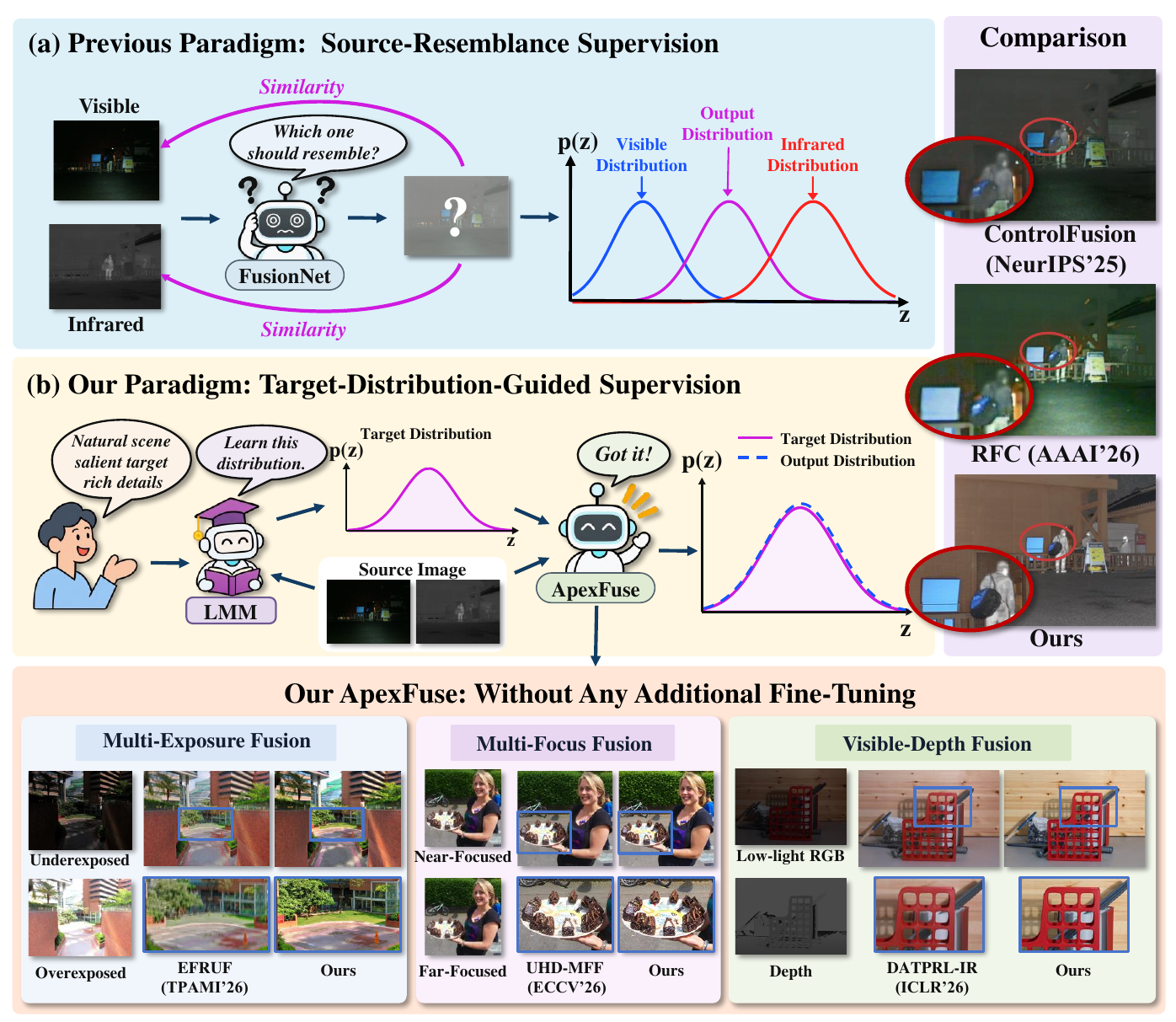}
	\caption{Illustration of our ApexFuse paradigm. (a) Previous fusion methods impose source-resemblance constraints that may conflict across modalities. (b) ApexFuse moves beyond source resemblance by learning a target distribution of high-quality scene representations. ApexFuse achieves SOTA performance in infrared-visible fusion and generalizes to multi-exposure, multi-focus, and visible-depth fusion without any additional fine-tuning.}
	\label{fig:intro}
\end{figure}

Unlike conventional image restoration tasks~\cite{yi2026ll}, image fusion lacks a naturally available ground truth, as no single sensor can directly capture the desired combination of multimodal characteristics. To circumvent this difficulty, existing approaches have largely defined fusion quality through information preservation~\cite{U2Fusion,zhang2026dpafuse}, encouraging the fused image to resemble its sources in terms of intensity, gradients, \textit{etc}., as illustrated in Fig.~\ref{fig:intro} (a). Yet, this seemingly intuitive principle conceals a fundamental paradox: \textit{How can a single image simultaneously resemble multiple observations whose appearances are inherently different}? Infrared and visible images are complementary precisely because they capture distinct physical properties. Therefore, enforcing their simultaneous resemblance may introduce competing demands, potentially compromising the unique strengths of each modality. More fundamentally, \textit{resemblance to source images does not necessarily imply an ideal integration of their complementary information}.

We argue that image fusion should move beyond source resemblance toward a more fundamental objective: \textit{faithfully realizing the complementary potential of multimodal observations}. An ideal fused image should not merely imitate its inputs, but reconstruct a visually natural and perceptually effective representation of the observed scene, preserving salient information and complementary details across modalities. In other words, source images should provide the factual foundation for reconstruction, rather than dictate the final visual appearance. From this perspective, image fusion can be viewed as a conditional scene reconstruction process, in which diverse source observations guide the recovery of high-quality scene representations. Importantly, this objective extends beyond specific modality combinations, reflecting a general pursuit of visual quality shared across diverse imaging tasks. This raises a compelling question: \textit{Can we explicitly define what an ideal fused image should look like, rather than merely constraining what it should resemble}?

Recent advances in large multimodal generative models (LMMs) offer a remarkable opportunity to bridge the long-standing gap between ideal fusion objectives and their visual realization. With powerful visual understanding, natural-language comprehension, and image generation capabilities~\cite{Emu3,VisionBanana}, these models can translate the desired characteristics of an ideal fused image into concrete visual references. Importantly, these characteristics are not arbitrary human preferences, but well-established goals of image fusion: preserving salient information from different modalities while presenting it in a coherent and natural manner. For instance, an ideal infrared-visible fused image should retain prominent thermal targets alongside the natural colors and fine textures of the visible scene, without introducing information unsupported by the original observations. What was once an abstract expectation can now be articulated in human language and translated into visual form through generative intelligence. This creates an unprecedented opportunity to move beyond indirectly defining fusion quality through source resemblance and make the desired fusion outcome explicit and learnable. More excitingly, it opens up the possibility of characterizing and learning a general distribution of high-quality scene representations, offering a generative objective that extends beyond individual image fusion tasks.

Motivated by this insight, we introduce ApexFuse, a new paradigm that turns the long-envisioned ideal of multimodal image fusion into a tangible and learnable reality, as shown in Fig.~\ref{fig:intro} (b). Rather than enforcing source resemblance, ApexFuse explicitly establishes what an ideal fused image should look like and learns to realize it. Specifically, we harness the powerful understanding and generation capabilities of LMMs to transform established fusion goals, expressed in natural language, into concrete visual references from infrared-visible image pairs. To ensure the reliability of these references, we develop a Human-LMM-in-the-Loop pipeline, in which LMMs produce candidate visual references that undergo iterative human evaluation, feedback-based refinement, and expert verification to identify and eliminate hallucinated or unsupported content. These carefully curated references provide direct supervision for target-distribution-guided adaptation of a pretrained FLUX model, transferring its rich generative priors and broad visual knowledge to high-quality multimodal scene reconstruction. This synergy empowers ApexFuse to produce remarkably clear, natural, and visually coherent fused images, seamlessly integrating salient thermal targets with vivid colors and intricate scene details, while achieving superior performance in downstream semantic segmentation and object detection. More importantly, this target-distribution-guided adaptation enables ApexFuse to learn a general distribution of high-quality scene representations, moving beyond modality-specific fusion mappings toward a unified generative image fusion paradigm. Remarkably, \textit{ApexFuse, trained solely for infrared-visible image fusion, can be directly applied to diverse imaging tasks without additional fine-tuning, such as multi-exposure fusion for high dynamic range (HDR)-like imaging, multi-focus fusion for extended depth-of-field (EDOF) imaging, and visible-depth fusion for low-light image enhancement (LLIE)}. We hope ApexFuse offers a new perspective on image fusion beyond source resemblance, enabling faithful multimodal integration and high-quality visual reconstruction.

\begin{figure*}[t]
    \centering
    \includegraphics[width=\textwidth]{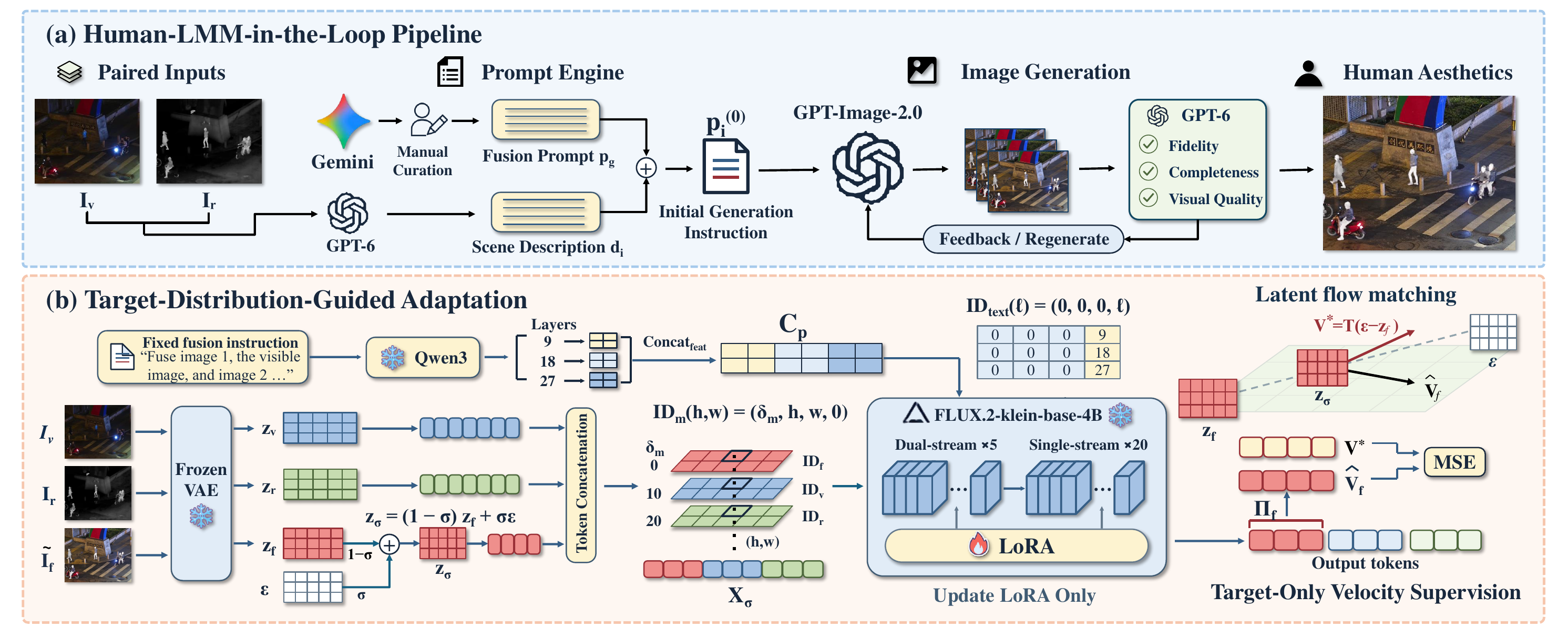}
    \caption{Overall Framework of our proposed ApexFuse.}
    \label{fig:method_overview}
\end{figure*}

In summary, our contributions are as follows:
\begin{itemize}
	\item We introduce ApexFuse, a novel generative image fusion paradigm that shifts the objective from source resemblance toward learning a general distribution of high-quality scene representations, enabling natural and coherent scene reconstruction beyond previous paradigms.
	
	\item We establish a human-LMM-in-the-loop pipeline to characterize the target distribution of high-quality scene representations, leveraging LMMs and iterative human feedback to transform abstract fusion objectives into reliable visual references.
	
    \item We transfer the generative priors of pretrained FLUX to image fusion through target-distribution-guided adaptation, achieving superior visual quality and downstream semantic performance over source-resemblance methods while harmoniously integrating salient thermal targets, natural colors, and fine textures.	
	
	\item ApexFuse demonstrates strong cross-task generalization, extending from infrared-visible fusion to multi-exposure fusion, multi-focus fusion, and visible-depth fusion without additional fine-tuning.
\end{itemize}

%% file: sec/2_methodology.tex
\section{Methodology}\label{sec:methodology}

\subsection{Problem Formulation}
ApexFuse aims to turn the long-envisioned ideal of multimodal image fusion into a tangible and learnable reality, and the overall framework is illustrated in Fig.~\ref{fig:method_overview}. Given a spatially aligned pair of visible and infrared images $[I_v,I_r]$, our goal is not to make the fused image resemble both sources in appearance. Instead, we seek to learn a general distribution of high-quality scene representations while preserving the factual scene information conveyed by the source images. To this end, ApexFuse consists of two stages.

\noindent\textbf{Stage 1}: We first establish a human-LMM-in-the-loop pipeline using LMMs and iterative human feedback, as shown in Fig.~\ref{fig:method_overview} (a). This pipeline translates abstract fusion objectives into reliable visual references, characterizing the target distribution of high-quality scene representations:
\begin{equation}
\tilde{I}_f =
\mathcal{R}\bigl(
\mathcal{G}_T(I_v,I_r,p);I_v,I_r
\bigr),
\label{eq:ideal_target}
\end{equation}
where $\tilde{I}_f$ denotes the ideal fusion target, $\mathcal{G}_T$ denotes a pretrained LMM for multimodal image generation, $p$ represents the fusion generation instruction, and $\mathcal{R}$ denotes the quality control process comprising iterative evaluation, feedback-based refinement, and human selection. Using these verified visual instances, we construct a supervised dataset of source image pairs and their corresponding ideal fusion targets:
\begin{equation}
\mathcal{D} =
\left\{
\left(
I_v^{(i)}, I_r^{(i)}, \tilde{I}_f^{(i)}
\right)
\right\}_{i=1}^{N}.
\label{eq:training_dataset}
\end{equation}

\noindent\textbf{Stage 2}: We transfer the generative priors of pretrained FLUX to image fusion through target-distribution-guided adaptation. Specifically, we adopt pretrained FLUX.2 as the generative backbone, freeze its original parameters $\theta_0$, and optimize only the LoRA adaptation parameters $\phi$. The model is trained using the following conditional flow matching objective:
\begin{equation}
\phi^* =
\arg\min_{\phi}
\mathbb{E}_{\mathcal{D}}
\left[
\mathcal{L}_{\mathrm{FM}}
\left(
\theta_0,\phi;
I_v,I_r,\tilde{I}_f
\right)
\right],
\label{eq:optimization_objective}
\end{equation}
where $\mathcal{L}_{\mathrm{FM}}$ denotes the conditional flow matching loss. Through these two stages, ApexFuse connects the construction of ideal fusion targets with the adaptation of pretrained generative priors, enabling the model to learn the target distribution of high-quality scene representations and realize it through image generation.

\subsection{Human-LMM-in-the-Loop Pipeline}
To characterize the target distribution of high-quality scene representations, we develop a human-LMM-in-the-loop pipeline that translates abstract fusion objectives into reliable visual references via prompt engine, image generation, and human aesthetics.

\noindent\textbf{Prompt Engine.}
We first employ Gemini to generate an initial prompt describing the ideal fusion objectives, which is subsequently refined through manual curation into a unified fusion prompt $p_g$ applicable across different scenes. To incorporate scene-specific information, we further employ GPT-6 to analyze each visible-infrared image pair and generate its corresponding scene description $d_i$. Then, the scene description is combined with the unified fusion prompt to formulate an initial generation instruction for each image pair, ensuring consistency in fusion objectives while accommodating scene-specific characteristics:
\begin{equation}
p_i^{(0)} = p_g \oplus d_i,
\label{eq:prompt_construction}
\end{equation}
where $\oplus$ denotes the combination of textual information.

\noindent\textbf{Image Generation.}
Given the $i$-th source image pair $[I_v^{(i)}, I_r^{(i)}]$ and its initial generation instruction $p_i^{(0)}$, we employ GPT-Image-2.0 to generate three candidate fused images, forming the initial candidate set:
\begin{equation}
\begin{aligned}
\mathcal{C}_i^{(0)}
&= \left\{ I_{i,j}^{(0)} \right\}_{j=1}^{3},\\
I_{i,j}^{(0)}
&= \left[
\mathcal{G}_T
\left(
I_v^{(i)}, I_r^{(i)}, p_i^{(0)}
\right)
\right]_j,
\end{aligned}
\label{eq:initial_candidates}
\end{equation}
where $j$ denotes the candidate image index, and $\mathcal{C}_i^{(0)}$ represents the initial candidate set. We then employ GPT to evaluate each candidate fused image against its corresponding visible and infrared source images. The evaluation considers three dimensions: \textit{Content Fidelity}, which examines whether the generated image introduces objects or scene content unsupported by the source images; \textit{Key Information Completeness}, which assesses whether salient targets or important scene information from the sources are missing; and \textit{Visual Quality}, which evaluates the naturalness, clarity, and overall visual coherence of the generated image.

Candidates that pass the evaluation are retained. For those that fail, we formulate individual feedback $r_{i,j}^{(t)}$ based on the evaluation results and corresponding refinement suggestions. The generation instruction for each candidate is then updated accordingly, followed by the regeneration of a single fused image:
\begin{equation}
\begin{aligned}
p_{i,j}^{(t+1)}
&= p_{i,j}^{(t)} \oplus r_{i,j}^{(t)},\\
I_{i,j}^{(t+1)}
&= \mathcal{G}_T\left(
I_v^{(i)}, I_r^{(i)}, p_{i,j}^{(t+1)}
\right),
\end{aligned}
\label{eq:iterative_refinement}
\end{equation}
here $p_{i,j}^{(0)}=p_i^{(0)}$, and $t$ denotes the refinement iteration index for each candidate. Each candidate maintains an independent feedback history and generation instruction. In subsequent iterations, only candidates that fail the evaluation undergo further refinement, with each candidate regenerated individually to produce a single image per iteration. The regenerated images are evaluated against the same criteria, and the process continues until all candidates meet the predefined quality requirements.

\noindent\textbf{Human Aesthetics.}
We conduct human aesthetic assessment of candidate fused images that pass the automated evaluation. Through qualitative comparison, we examine their content consistency with the source images and overall visual appeal. The final ideal fusion target $\widetilde{I}_f$ is then selected from the qualified candidates for subsequent supervised learning of the fusion model.

\subsection{Target-Distribution-Guided Adaptation}
High-quality ideal fusion outcomes provide explicit visual supervision, but how to transfer the rich prior knowledge of pretrained generative models to the image fusion task remains a key challenge. To this end, we adopt the pretrained FLUX.2-klein-base-4B~\cite{FLUX2} as our generative backbone, jointly condition the generation process on visible images, infrared images, and a fixed fusion instruction. We further employ Low-Rank Adaptation (LoRA) to transfer generative priors through target-distribution-guided adaptation, enabling the model to learn the conditional generation of ideal fusion outcomes.

\noindent\textbf{Image-Pair and Text Conditioning.}
To condition the pretrained FLUX model on both visible and infrared images and a fixed fusion instruction, we construct image-reference and text conditions separately. Given a paired visible and infrared images $[I_v,I_r]$, and their corresponding ideal fusion target $\widetilde{I}_f$ during training, we first encode all three images into the latent space using a shared, frozen VAE encoder. The resulting latent representations are then spatially rearranged and normalized:
\begin{equation}
z_m =
\mathcal{N}\left(
\mathcal{P}\left(\mathcal{E}(I_m)\right)
\right),
\quad I_m\in\{I_v,I_r,\widetilde{I}_f\},
\label{eq:latent_representation}
\end{equation}
where $\mathcal{E}$ denotes the VAE encoding operation, $\mathcal{P}$ rearranges non-overlapping $2\times2$ spatial patches into the channel dimension, and $\mathcal{N}$ denotes normalization based on the pretrained VAE statistics.

Then, the visible and infrared latents $z_v$ and $z_r$ serve as unperturbed reference conditions. They are tokenized and concatenated with the noisy target latent $z_\sigma$ along the sequence dimension:
\begin{equation}
X_\sigma =
\left[
\mathcal{T}(z_\sigma);
\mathcal{T}(z_v);
\mathcal{T}(z_r)
\right],
\label{eq:image_token_concatenation}
\end{equation}
where $\mathcal{T}$ denotes the operation of flattening the latent representations along the spatial dimensions into token sequences, and $z_\sigma$ denotes the noisy latent of the fusion target.

To distinguish the fusion target from the two reference images while preserving their respective spatial information, we construct four-dimensional positional identifiers for the three groups of image tokens:
\begin{equation}
\begin{aligned}
\mathrm{ID}_m(h,w)
&=(\delta_m,h,w,0),
\quad m\in\{f,v,r\},\\
(\delta_f,\delta_v,\delta_r)
&=(0,10,20),
\end{aligned}
\label{eq:image_position_ids}
\end{equation}
where $(h,w)$ denotes the spatial coordinates of an image token, and $\delta_m$ represents the positional offset assigned to each image. These positional identifiers are incorporated into attention computation through the multi-axis Rotary Position Embedding (RoPE) mechanism of FLUX.

For text conditioning, we employ a frozen Qwen3 text encoder to extract token-level semantic representations from the fixed fusion instruction. Specifically, we extract hidden features from the 9th, 18th, and 27th layers of the text encoder and concatenate them along the feature dimension to obtain the text condition:
\begin{equation}
C_p =
\operatorname{Concat}_{\mathrm{feat}}
\left(
H_p^{(9)},H_p^{(18)},H_p^{(27)}
\right),
\label{eq:text_condition}
\end{equation}
here $H_p^{(l)}$ denotes the hidden features from the $l$-th layer of the text encoder. For the $\ell$-th text token, its four-dimensional positional identifier is defined as:
\begin{equation}
\mathrm{ID}_{\mathrm{text}}(\ell)=(0,0,0,\ell).
\label{eq:text_position_ids}
\end{equation}
The corresponding positional information is likewise incorporated into attention computation via multi-axis RoPE.

Finally, the image token sequence $X_\sigma$ and text features $C_p$ are jointly fed into the FLUX Transformer. Through joint attention in the dual-stream blocks and unified processing in the single-stream blocks, the model facilitates information exchange among the fusion target, multimodal reference images, and text conditions.

\noindent\textbf{LoRA-Based FLUX Fine-Tuning.}
To transfer the generative and editing capabilities of pretrained FLUX to the multimodal image fusion task, we freeze its original parameters $\theta_0$ and introduce trainable Low-Rank Adaptation (LoRA) parameters $\phi$. These parameters are incorporated into the image-side attention projection layers of the dual-stream blocks, as well as the attention and MLP projection layers of single-stream blocks. For each adapted pretrained weight matrix $W_0$, the effective weight is defined as:
\begin{equation}
W = W_0 + \frac{\alpha}{r}BA,
\label{eq:lora_adaptation}
\end{equation}
where $A$ and $B$ are trainable low-rank matrices, $r$ denotes the adaptation rank, and $\alpha$ is the scaling factor.

During fine-tuning, we first sample timestep indices using a logit-normal sampling strategy and retrieve the corresponding noise level $\sigma$ and timestep $t_\sigma$ from a predefined scheduler. We sample standard Gaussian noise $\varepsilon$ and construct a linear interpolation path between the ideal fusion target latent $z_f$ and the noise:
\begin{equation}
z_\sigma = (1-\sigma)z_f + \sigma\varepsilon,
\quad
\varepsilon \sim \mathcal{N}(0,I),
\label{eq:flow_interpolation}
\end{equation}
where $\sigma\in[0,1]$ denotes the noise level. Differentiating this interpolation path with respect to $\sigma$ yields the corresponding target velocity:
\begin{equation}
v^* = \frac{\partial z_\sigma}{\partial \sigma}
= \varepsilon-z_f.
\label{eq:target_velocity}
\end{equation}

Subsequently, the constructed image token sequence $X_\sigma$, timestep $t_\sigma$, text condition $C_p$, and corresponding positional identifiers are fed into the LoRA-adapted FLUX Transformer to predict the target velocity $\widehat{V}_f$. The prediction is supervised using the mean squared error between the predicted and ground-truth velocities. Accordingly, the conditional flow matching loss is formulated as:
\begin{equation}
\begin{gathered}
\widehat{V}_f =
\Pi_f F_{\theta_0,\phi}
(X_\sigma,t_\sigma,C_p;\mathrm{IDs}),\\[3pt]
\mathcal{L}_{\mathrm{FM}}(\phi)
=
\mathbb{E}_{\mathcal{D},k,\varepsilon,b}
\left[
\frac{1}{N_zd_z}
\left\|
\widehat{V}_f-\mathcal{T}(\varepsilon-z_f)
\right\|_F^2
\right],
\end{gathered}
\label{eq:flow_matching_loss}
\end{equation}
where $F_{\theta_0,\phi}$ denotes the LoRA-adapted FLUX Transformer, $\Pi_f$ denotes the operation of extracting the output tokens corresponding to the fusion target, $N_z$ and $d_z$ represent the number of target tokens and their feature dimensionality, respectively, and $\mathrm{IDs}$ denotes the positional identifiers of the image and text tokens. By minimizing this loss, gradients are backpropagated to update only the LoRA parameters $\phi$. This enables the model to learn the velocity field for generating ideal fusion outcomes conditioned on both the source image pair and the text instruction.

During inference, given a visible–infrared image pair and a fixed default fusion instruction, we encode both images into latent representations using the frozen VAE encoder and initialize the fusion latent with Gaussian noise. The fine-tuned FLUX model then progressively denoises the latent through iterative flow-matching sampling, conditioned on the source images and text instruction. After inverse normalization and spatial rearrangement, the generated latent is decoded by the frozen VAE decoder to obtain the fused image. Notably, ApexFuse learns a general target distribution of high-quality scene representations rather than a modality-specific fusion mapping. Therefore, its generative priors are not restricted to infrared–visible fusion, enabling generalization to diverse fusion tasks without additional fine-tuning.

\begin{figure*}[t]
	\centering
	\includegraphics[width=\textwidth]{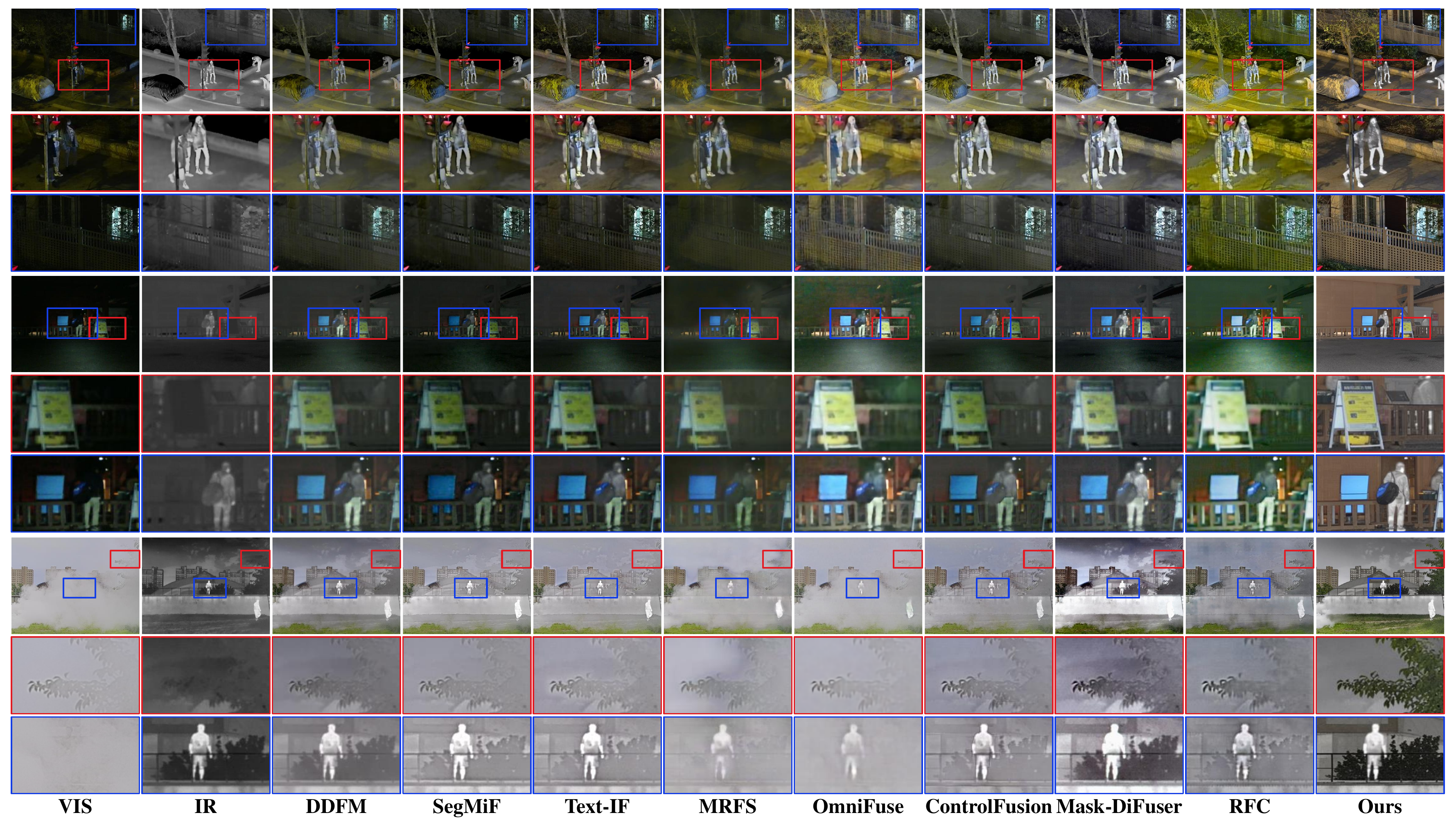}
	\vspace{-0.7cm}
	\caption{Qualitative comparisons of visual representation capability on the LLVIP, MFNet, and FMB datasets.}
	\label{fig:Fusion_Comparison}
\end{figure*}

%% file: sec/3_experiments.tex
\section{Experiments}

\subsection{Experimental Settings}
\noindent\textbf{Datasets.}
We select $1,060$ infrared-visible image pairs from LLVIP~\cite{LLVIP}, and generate ideal fusion targets using GPT-Image-2.0. The data are randomly split within each scene into $810$ training pairs and $250$ testing pairs.

\noindent\textbf{Implementation Details.}
We employ LoRA for fine-tuning, with a rank $r=32$ and scaling factor $\alpha=32$. Training runs for $3,000$ steps using AdamW, with an initial learning rate of $5\times10^{-5}$ and weight decay of $0.01$. During training, We adopt a cosine learning rate schedule with $100$ warmup steps and a gradient clipping threshold of $1.0$. All experiments are conducted on NVIDIA GeForce RTX 4090 GPUs with an Intel Xeon Platinum 8368Q CPU.

\begin{table*}[t]
    \centering
    \caption{comparisons of visual representation capability on the LLVIP, MFNet, and FMB datasets.}
    \label{tab:visual_comparison}
    \vspace{-0.3cm}
    \scriptsize
    \setlength{\tabcolsep}{2.1pt}
    \renewcommand{\arraystretch}{1}

    \begin{tabular}{@{}c@{\hspace{5pt}}cccc
                      @{\hspace{7pt}}cccc
                      @{\hspace{7pt}}cccc@{}}
        \toprule
        & \multicolumn{4}{c}{\textbf{LLVIP}}
        & \multicolumn{4}{c}{\textbf{MFNet}}
        & \multicolumn{4}{c}{\textbf{FMB}} \\
        \cmidrule(lr){2-5}
        \cmidrule(lr){6-9}
        \cmidrule(lr){10-13}

        \textbf{Method}
        & \textbf{MUSIQ}$\uparrow$
        & \textbf{TReS}$\uparrow$
        & \textbf{MANIQA}$\uparrow$
        & \textbf{CLIP-IQA}$\uparrow$

        & \textbf{MUSIQ}$\uparrow$
        & \textbf{TReS}$\uparrow$
        & \textbf{MANIQA}$\uparrow$
        & \textbf{CLIP-IQA}$\uparrow$

        & \textbf{MUSIQ}$\uparrow$
        & \textbf{TReS}$\uparrow$
        & \textbf{MANIQA}$\uparrow$
        & \textbf{CLIP-IQA}$\uparrow$ \\
        \midrule

        DDFM
        & 59.379 & 77.216 & 0.351 & 0.319
        & 37.794 & 34.964 & 0.167 & 0.144
        & 52.110 & 56.638
        & \cellcolor{secondcolor}\textbf{0.279}
        & 0.199 \\

        SegMiF
        & 62.241 & 78.120 & 0.361 & 0.324
        & 37.046 & 33.824 & 0.151 & 0.135
        & 54.127 & 61.539 & 0.262 & 0.183 \\

        Text-IF
        & 62.335 & 81.153 & 0.355 & 0.370
        & 40.464 & 36.940 & 0.163 & 0.165
        & 54.704 & 63.183 & 0.259 & 0.160 \\

        MRFS
        & 57.198 & 70.338 & 0.300 & 0.247
        & 36.358 & 34.606 & 0.147 & 0.131
        & 53.083 & 58.585 & 0.252 & 0.184 \\

        OmniFuse
        & 54.425 & 55.154 & 0.266 & 0.315
        & 35.305 & 32.017 & 0.154
        & \cellcolor{secondcolor}\textbf{0.214}
        & 50.684 & 49.139 & 0.266
        & \cellcolor{secondcolor}\textbf{0.222} \\

        ControlFusion
        & 64.179 & 84.224
        & \cellcolor{secondcolor}\textbf{0.373}
        & 0.373

        & \cellcolor{secondcolor}\textbf{42.022}
        & \cellcolor{secondcolor}\textbf{41.545}
        & \cellcolor{secondcolor}\textbf{0.168}
        & 0.145

        & \cellcolor{secondcolor}\textbf{56.341}
        & \cellcolor{secondcolor}\textbf{65.925}
        & 0.277 & 0.191 \\

        Mask-DiFuser
        & 61.138 & 76.640 & 0.344 & 0.368
        & 37.342 & 29.384 & 0.140 & 0.128
        & 47.822 & 54.199 & 0.210 & 0.189 \\

        RFC
        & \cellcolor{secondcolor}\textbf{65.332}
        & \cellcolor{secondcolor}\textbf{84.307}
        & 0.334
        & \cellcolor{secondcolor}\textbf{0.411}

        & 36.610 & 38.811 & 0.147 & 0.164
        & 53.541 & 62.720 & 0.251 & 0.216 \\

        \midrule

        \rowcolor{bestcolor}
        \textbf{Ours}
        & \textbf{70.206}
        & \textbf{91.391}
        & \textbf{0.537}
        & \textbf{0.564}

        & \textbf{64.927}
        & \textbf{81.998}
        & \textbf{0.393}
        & \textbf{0.463}

        & \textbf{68.630}
        & \textbf{83.160}
        & \textbf{0.424}
        & \textbf{0.476} \\

        \bottomrule
    \end{tabular}
\end{table*}

\subsection{Visual Representation Comparison}
We compare the visual representation capability of ApexFuse with eight SOTA fusion methods: DDFM~\cite{DDFM}, SegMiF~\cite{SegMiF}, Text-IF~\cite{TextIF}, MRFS~\cite{MRFS}, OmniFuse~\cite{OmniFuse}, ControlFusion~\cite{ControlFusion}, Mask-DiFuser~\cite{MaskDiFuser}, and RFC~\cite{RFC}, on LLVIP~\cite{LLVIP}, FMB~\cite{SegMiF}, and MFNet~\cite{MFNet} datasets. ApexFuse is trained exclusively on LLVIP and directly evaluated on FMB and MFNet to assess cross-dataset generalization. As shown in Fig.~\ref{fig:Fusion_Comparison}, ApexFuse delivers a \textbf{\textit{strikingly distinct visual experience, producing exceptionally clear, natural, and visually compelling results}}. It faithfully preserves thermal targets and intricate visible structures, while remarkably recovering rich scene details that are severely degraded or barely discernible in the source images.

The quantitative results in Table~\ref{tab:visual_comparison} are evaluated using four image quality metrics: MUSIQ~\cite{MUSIQ}, TReS~\cite{TReS}, MANIQA~\cite{MANIQA}, and CLIP-IQA~\cite{CLIPIQA}. ApexFuse achieves the best performance across all four metrics on all three datasets, demonstrating superior visual quality and cross-dataset generalization. These results further validate its ability to move beyond source resemblance toward high-quality visual reconstruction.

\subsection{Semantic Representation Comparison}
\begin{figure}[t]
    \centering
    \includegraphics[width=\columnwidth]{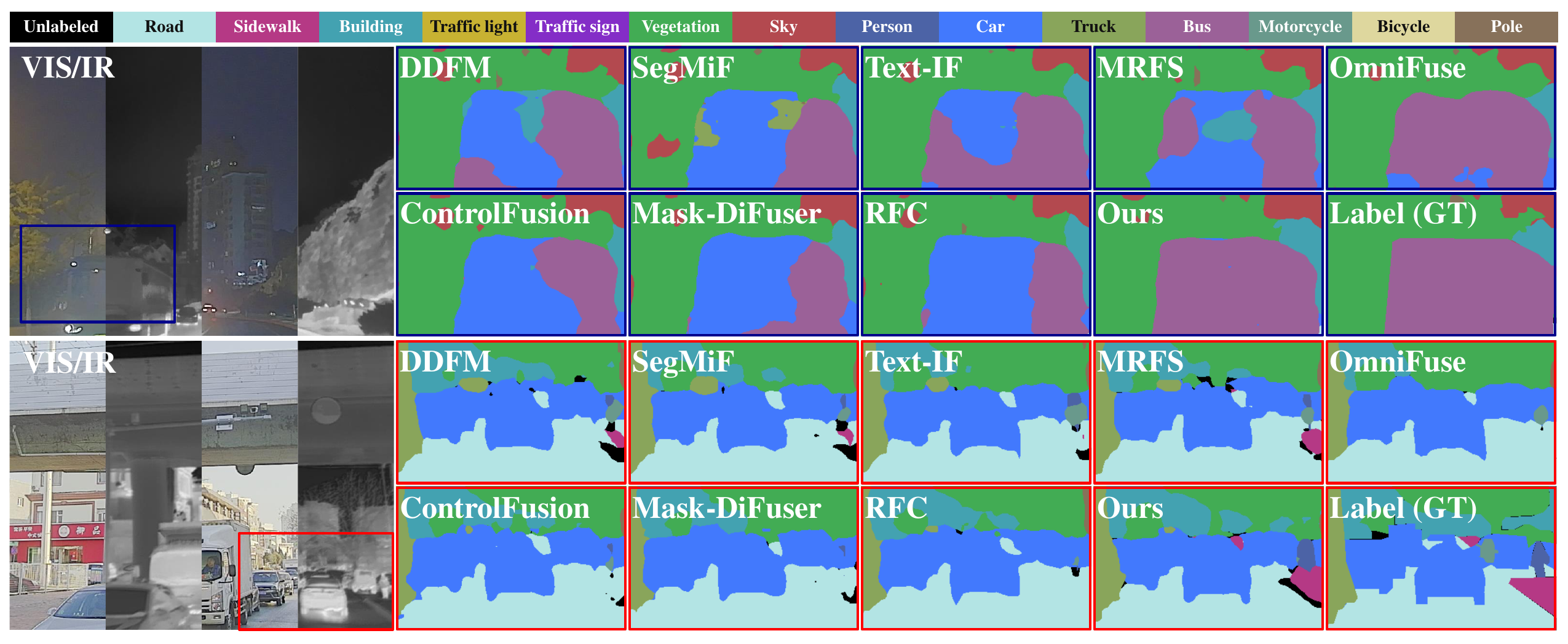}
    \vspace{-0.7cm}
    \caption{Qualitative results of semantic segmentation.}
    \label{fig:Segmentation}
    \vspace{-0.2cm}
\end{figure}

\noindent\textbf{Semantic Segmentation.}
To further evaluate the scene-level semantic representation capability of ApexFuse, we conduct semantic segmentation experiments on the FMB dataset. Specifically, we independently retrain SegNeXt~\cite{SegNeXt} using the fused images generated by each method and evaluate their performance under identical testing settings. As shown in Fig.~\ref{fig:Segmentation}, ApexFuse produces clearer scene structures, leading to more accurate segmentation, particularly in object completeness and semantic region delineation. The quantitative results in Table~\ref{tab:segmentation_comparison} further demonstrate the superior semantic representation capability of ApexFuse. These results confirm that ApexFuse provides more reliable scene information for downstream visual understanding while maintaining high visual quality.

\begin{table}[t]
    \centering
    \caption{Quantitative comparison of semantic segmentation.}
    \vspace{-0.3cm}
    \label{tab:segmentation_comparison}
    \scriptsize
    \setlength{\tabcolsep}{1.8pt}
    \renewcommand{\arraystretch}{1}

    \begin{tabular}{@{}c|cccccccc|c@{}}
        \toprule
        \textbf{Method}
        & \textbf{Car}
        & \textbf{Pers.}
        & \textbf{Bus}
        & \textbf{Build.}
        & \textbf{Lamp}
        & \textbf{Sign}
        & \textbf{Road}
        & \textbf{Side.}
        & \textbf{mIoU}$\uparrow$ \\
        \midrule

        DDFM
        & 79.67 & 67.77 & 47.31 & 83.75
        & 45.14 & 73.95 & 87.44 & 47.55
        & 57.84 \\

        SegMiF
        & 80.49 & 68.85 & 49.66 & 83.01
        & 34.90 & 71.20 & 86.74 & 46.36
        & 57.29 \\

        \rowcolor{secondcolor}
        Text-IF
        & 79.95 & 69.12 & 55.61 & 83.68
        & 40.20 & 72.56 & 87.67 & 49.47
        & \textbf{58.54} \\

        MRFS
        & 79.79 & 68.45 & 57.27 & 83.04
        & 40.06 & 71.90 & 87.10 & 46.83
        & 57.84 \\

        OmniFuse
        & 79.86 & 66.24 & 47.70 & 80.11
        & 36.66 & 72.09 & 86.22 & 46.53
        & 55.94 \\

        ControlFusion
        & 79.57 & 69.82 & 53.52 & 83.99
        & 41.10 & 73.23 & 87.52 & 48.73
        & 58.41 \\

        Mask-DiFuser
        & 79.42 & 69.07 & 50.55 & 83.39
        & 34.85 & 68.73 & 87.17 & 48.23
        & 56.78 \\

        RFC
        & 80.23 & 69.50 & 65.24 & 83.02
        & 38.74 & 68.36 & 86.25 & 44.57
        & 57.70 \\

        \midrule

        \rowcolor{bestcolor}
        \textbf{Ours}
        & 81.05 & 67.72 & 64.41 & 83.84
        & 46.83 & 74.96 & 86.16 & 49.05
        & \textbf{59.05} \\

        \bottomrule
    \end{tabular}
    \vspace{-0.cm}
\end{table}

\begin{figure}[t]
	\centering
	\includegraphics[width=\columnwidth]{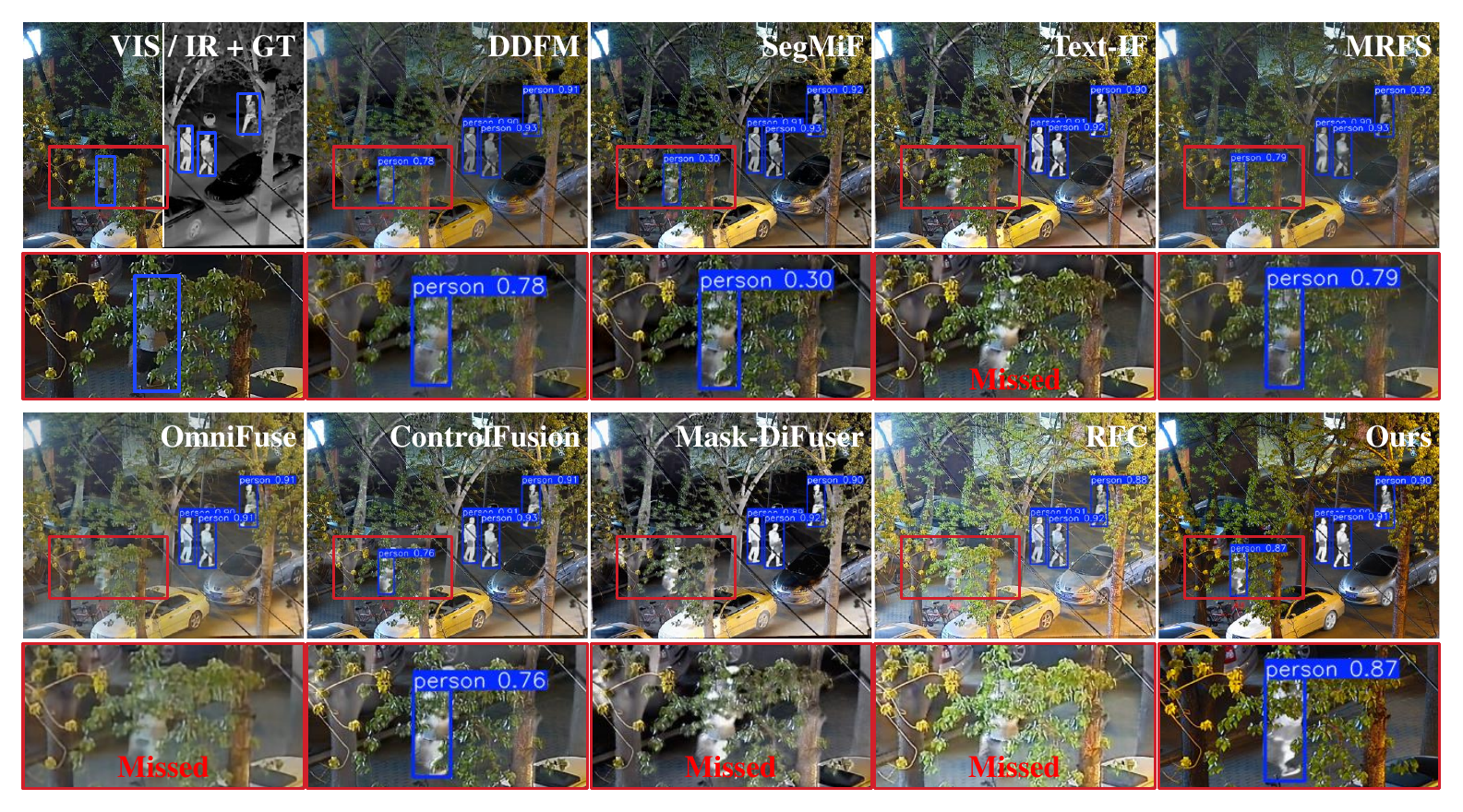}
	\vspace{-0.7cm}
	\caption{Qualitative results of object detection.}
	\label{fig:Detection}
	\vspace{-0.2cm}
\end{figure}

\begin{table}[t]
	\centering
	\caption{Quantitative comparison of object detection.}
	\vspace{-0.3cm}
	\label{tab:detection_comparison}
	\footnotesize
	\setlength{\tabcolsep}{1.8pt}
	\renewcommand{\arraystretch}{1}
	
	\begin{tabular}{@{}c|ccccc@{}}
		\toprule
		\textbf{Method}
		& \textbf{Precision}$\uparrow$
		& \textbf{Recall}$\uparrow$
		& \textbf{mAP}$_{50}$\,$\uparrow$
		& \textbf{mAP}$_{75}$\,$\uparrow$
		& \textbf{mAP}$_{50:95}$\,$\uparrow$ \\
		\midrule
		
		DDFM
		& 96.090 & 92.407 & 95.869
		& 71.701 & 62.832 \\
		
		SegMiF
		& 94.486 & 90.832 & 94.927
		& 71.979 & 62.981 \\
		
		Text-IF
		& 94.473 & 90.600 & 95.330
		& 70.843 & 63.539 \\
		
		MRFS
		& 96.414 & 91.547 & 96.048
		& 71.714 & 62.325 \\
		
		OmniFuse
		& 93.074 & 89.398 & 94.501
		& 68.071 & 59.856 \\
		
		ControlFusion
		& 95.464 & 91.547 & 95.882
		& 72.928 & 63.471 \\
		
		\rowcolor{secondcolor}
		Mask-DiFuser
		& 95.794 & 90.688 & 95.816
		& 74.182 & \textbf{63.872} \\
		
		RFC
		& 95.455 & 90.264 & 95.569
		& 68.368 & 61.440 \\
		
		\midrule
		
		\rowcolor{bestcolor}
		\textbf{Ours}
		& 94.500 & 92.693 & 96.271
		& 77.260 & \textbf{65.353} \\
		
		\bottomrule
	\end{tabular}
\end{table}

\noindent\textbf{Object Detection.}
We conduct object detection experiments on the LLVIP dataset. Specifically, we independently retrain YOLOv8 using the fused images generated by each method and evaluate its performance under identical testing settings. As shown in Fig.~\ref{fig:Detection}, ApexFuse successfully preserves target contours and structural details. This enables clear detection of all targets, including partially occluded ones, with the highest confidence scores. The quantitative results in Table~\ref{tab:detection_comparison} further demonstrate its superior target representation capability. These results confirm that ApexFuse provides reliable target information for downstream object detection while maintaining high visual quality.

\begin{figure}[t]
	\centering
	\includegraphics[width=\columnwidth]{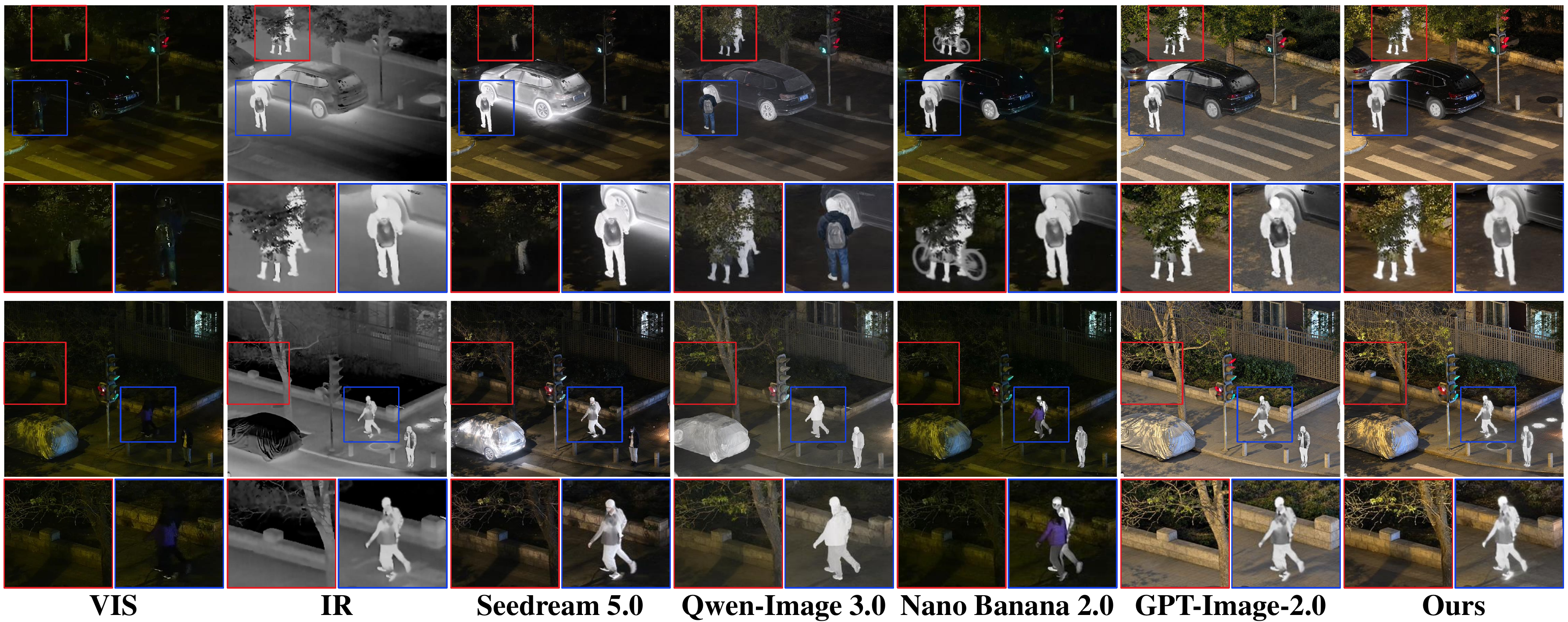}
	\vspace{-0.7cm}
	\caption{Qualitative comparison with commercial LMMs.}
	\label{fig:lmm_comparison}
	\vspace{-0.2cm}
\end{figure}

\begin{table}[t]
	\centering
	\caption{Quantitative comparison with commercial LMMs.}
	\vspace{-0.3cm}
	\label{tab:lmm_comparison}
	\footnotesize
	\setlength{\tabcolsep}{3pt}
	\renewcommand{\arraystretch}{1}
		\begin{tabular}{@{}c|cccc@{}}
			\toprule
			\textbf{Model}
			& \textbf{MUSIQ}$\uparrow$
			& \textbf{TReS}$\uparrow$
			& \textbf{MANIQA}$\uparrow$
			& \textbf{CLIP-IQA}$\uparrow$ \\
			\midrule
			
			Seedream 5.0
			& 65.151
			& 84.445
			& 0.448
			& \cellcolor{bestcolor}\textbf{0.571} \\
			
			Qwen-Image 3.0
			& 67.829
			& 87.167
			& 0.449
			& 0.492 \\
			
			Nano Banana 2.0
			& 62.444
			& 73.384
			& 0.361
			& 0.433 \\
			
			GPT-Image-2.0
			& \cellcolor{secondcolor}70.868
			& \cellcolor{secondcolor}91.620
			& \cellcolor{secondcolor}0.521
			& 0.489 \\
			
			\midrule
			
			\textbf{Ours}
			& \cellcolor{bestcolor}\textbf{71.434}
			& \cellcolor{bestcolor}\textbf{92.329}
			& \cellcolor{bestcolor}\textbf{0.546}
			& \cellcolor{secondcolor}0.571 \\
			
			\bottomrule
		\end{tabular}%
\end{table}

\subsection{Comparison with Commercial LMMs}
To investigate whether target-distribution-guided adaptation effectively transfers general-purpose generative priors to image fusion, we compare ApexFuse with four representative commercial LMMs: Seedream 5.0, Qwen-Image 3.0, Nano Banana 2.0, and GPT-Image-2.0. Due to API usage constraints, we conduct the comparison on $20$ randomly selected infrared-visible image pairs, providing all models with the same fusion prompt: “\textit{Please generate an enhanced infrared-visible fused image from the original$\ldots$}”.

As shown in Fig.~\ref{fig:lmm_comparison} and Table~\ref{tab:lmm_comparison}, ApexFuse achieves superior cross-modal information integration and fused image quality, validating the effectiveness of target-distribution-guided adaptation. Notably, although GPT-Image-2.0 is used to construct the fusion targets for ApexFuse, our model still outperforms it. This advantage stems from learning cross-modal integration patterns across diverse scenes and imaging conditions from numerous paired source images and high-quality fusion targets, going beyond the direct, instruction-driven generation of general-purpose models. Moreover, ApexFuse requires no external LMMs during inference, facilitating independent deployment.

\begin{figure}[t]
	\centering
	\includegraphics[width=\columnwidth]{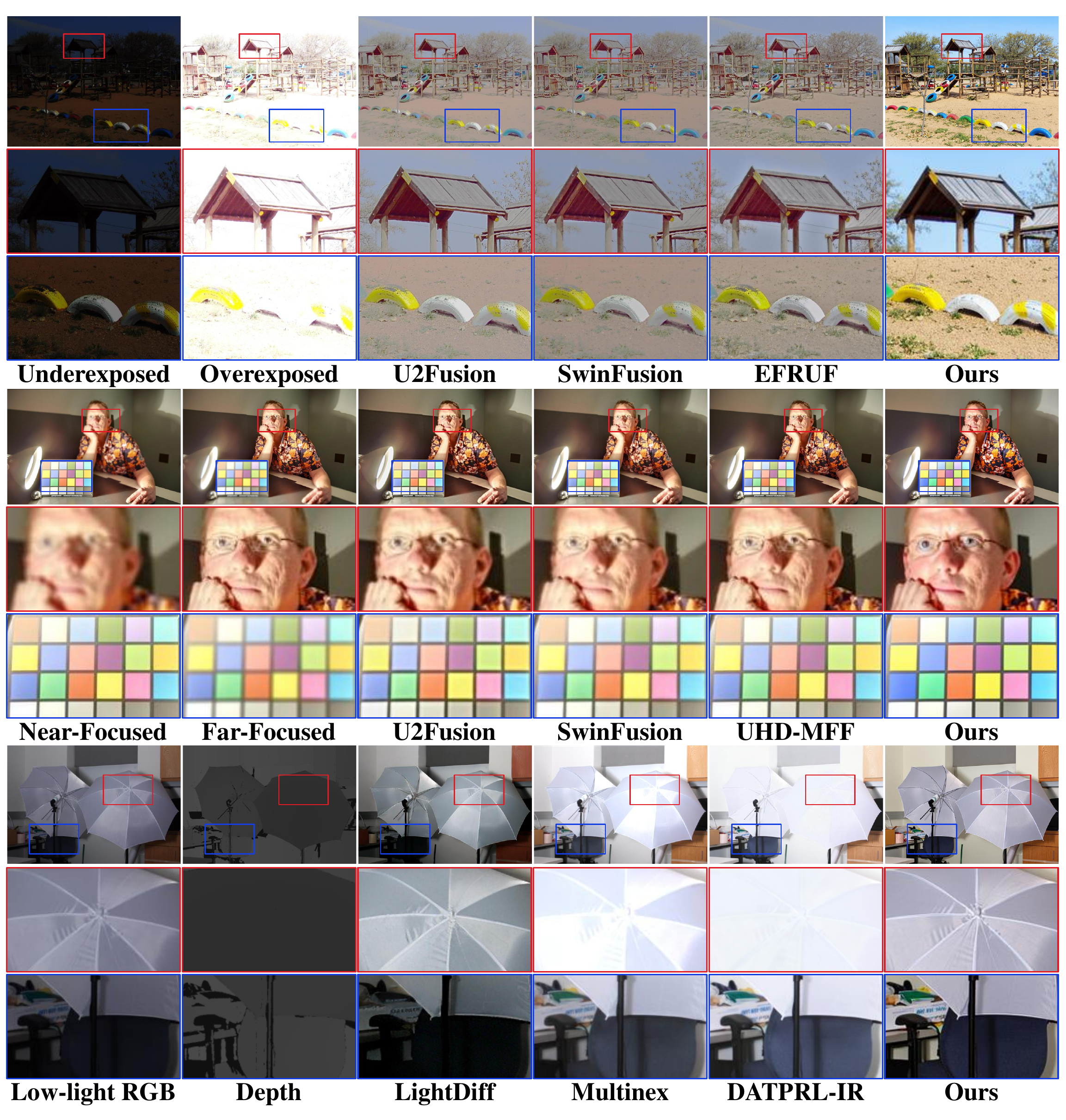}
	\vspace{-0.7cm}
	\caption{Qualitative comparison of cross-task generalization.}
	\label{fig:cross_task_qualitative}
\end{figure}

\begin{figure}[t]
	\centering
	\includegraphics[width=\columnwidth]{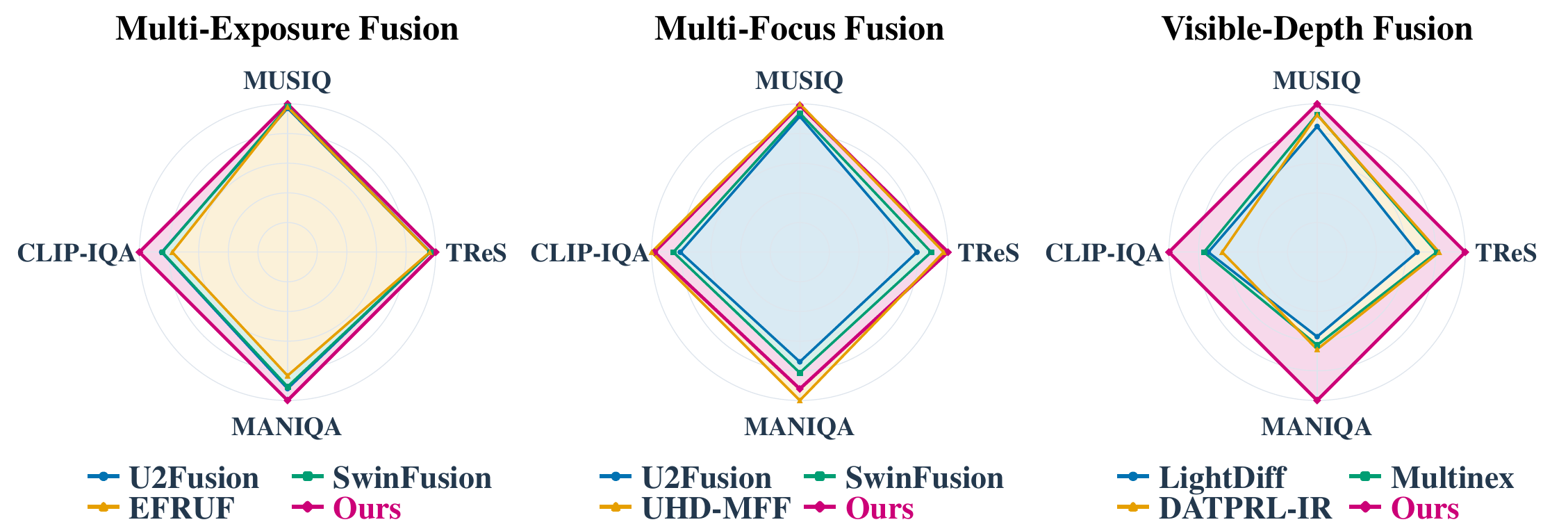}
	\vspace{-0.7cm}
	\caption{Quantitative comparison of cross-task generalization.}
	\label{fig:cross_task_quantitative}
\end{figure}

\subsection{Cross-Task Generalization without Fine-tuning}
To evaluate the cross-task generalization of ApexFuse, we directly apply the model trained exclusively on infrared-visible fusion to multi-exposure, multi-focus, and visible-depth fusion \textbf{\textit{without additional fine-tuning}}. We compare it with U2Fusion~\cite{U2Fusion}, SwinFusion~\cite{SwinFusion}, and EFRUF~\cite{EFRUF} for multi-exposure fusion; U2Fusion, SwinFusion, and UHD-MFF~\cite{UHDMFF} for multi-focus fusion; and LightDiff~\cite{LightDiff}, Multinex~\cite{Multinex}, and DATPRL-IR~\cite{DATPRLIR} for visible-depth fusion.

As shown in Fig.~\ref{fig:cross_task_qualitative}, remarkably, ApexFuse delivers consistently impressive visual quality across all three tasks despite being trained exclusively for infrared-visible fusion. It produces naturally exposed scenes with balanced illumination, uniformly sharp details across the entire scene, and vivid, natural colors with rich textures. More importantly, \textbf{\textit{ApexFuse goes beyond preserving source information, effectively inferring and reconstructing scene details that are degraded or barely discernible in the original observations}}. The quantitative results in Fig.~\ref{fig:cross_task_quantitative} further show that ApexFuse achieves superior performance across all three tasks. These results demonstrate that its learned generative priors extend beyond specific modality combinations, enabling high-quality scene reconstruction across diverse imaging tasks without additional fine-tuning.

\begin{figure}[t]
	\centering
	\includegraphics[width=\columnwidth]{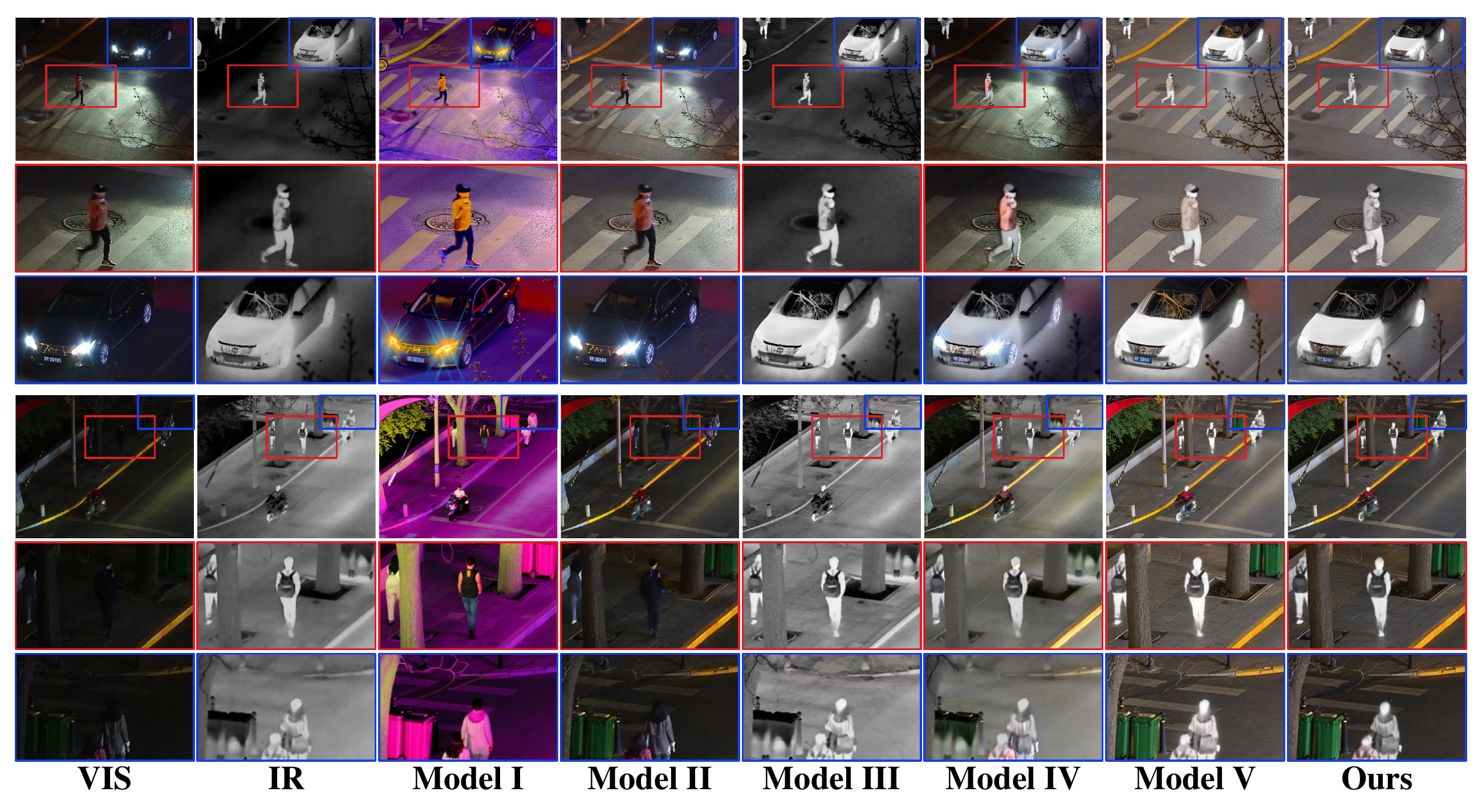}
	\vspace{-0.7cm}
	\caption{Qualitative comparison of different ablation variants.}
	\label{fig:ablation}
	\vspace{-0.2cm}
\end{figure}

\begin{table}[t]
	\centering
	\caption{Quantitative comparison of different ablation variants.}
	\vspace{-0.3cm}
	\label{tab:ablation}
	\footnotesize
	\setlength{\tabcolsep}{6.5pt}
	\renewcommand{\arraystretch}{1}
	\begin{tabular}{@{}c|cccc@{}}
		\toprule
		\textbf{Method}
		& \textbf{MUSIQ}$\uparrow$
		& \textbf{TReS}$\uparrow$
		& \textbf{MANIQA}$\uparrow$
		& \textbf{CLIP-IQA}$\uparrow$ \\
		\midrule
		
		Model I
		& 68.0981 & 83.7877 & 0.4820 & 0.5149 \\
		
		Model II
		& 64.1306 & 87.2940 & 0.4512 & 0.4793 \\
		
		Model III
		& 58.6494 & 65.2896 & 0.3743 & 0.4121 \\
		
		Model IV
		& 65.0726 & 86.9249 & 0.3910 & 0.4709 \\
		
		\rowcolor{bestcolor}
		\textbf{Model V}
		& \textbf{71.1664}
		& \textbf{91.8312}
		& \textbf{0.5520}
		& \textbf{0.5807} \\
		
		\midrule
		
		\rowcolor{secondcolor}
		\textbf{Ours}
		& \textbf{70.2061}
		& \textbf{91.3913}
		& \textbf{0.5372}
		& \textbf{0.5644} \\
		
		\bottomrule
	\end{tabular}
\end{table}

\subsection{Ablation Study}
We investigate four key aspects of ApexFuse: generative prior adaptation, cross-modal complementarity, supervision targets, and model scale. Five variants are constructed: \textbf{Model I} uses pretrained FLUX.2 4B without LoRA fine-tuning; \textbf{Models II and III} replace the visible-infrared inputs with two identical visible or infrared images, respectively; \textbf{Model IV} replaces the ideal fusion targets with the best of eight source-resemblance-based fusion candidates; and \textbf{Model V} replaces FLUX.2 4B with FLUX.2 9B. All variants are trained under otherwise identical settings.

Fig.~\ref{fig:ablation} and Table~\ref{tab:ablation} show the qualitative and quantitative results on the LLVIP dataset. Compared with Model I, ApexFuse produces higher-quality results, demonstrating the effectiveness of target-distribution-guided adaptation in transferring pretrained generative priors. Models II and III exhibit inferior results due to the loss of cross-modal complementarity, confirming the importance of multimodal fusion. Compared with Model IV, ApexFuse achieves superior visual quality and quantitative performance, validating the advantage of learning high-quality scene representations beyond source resemblance. Finally, Model V demonstrates the scalability of ApexFuse, with stronger generative priors yielding further improvements in fusion quality.

%% file: sec/4_conclusion.tex
\section{Conclusion}\label{sec:Conclusion}
We propose ApexFuse, a generative fusion paradigm that moves beyond source resemblance to learn a general distribution of high-quality scene representations. A human-LMM-in-the-loop pipeline constructs reliable visual references to characterize this target distribution, while the target-distribution-guided adaptation transfers pretrained FLUX generative priors to image fusion. Extensive experiments demonstrate superior visual quality and downstream semantic performance. Remarkably, ApexFuse generalizes to multi-exposure, multi-focus, and visible-depth fusion without additional fine-tuning.

%% file: sec/5_acknowledgments.tex
\section*{Acknowledgments}
This work was supported by the NSFC (62506268, 62676289), the Basic Research Program of Jiangsu (BK20250454), and the CPSF (GZB20250066, 2025M78150,2026T190426). The numerical calculations in this paper have been done on the supercomputing system in the Supercomputing Center of Wuhan University. The authors declare no competing interests.

%% file: ApexFuse.bbl
\begin{thebibliography}{29}
\providecommand{\natexlab}[1]{#1}
\providecommand{\url}[1]{\texttt{#1}}
\expandafter\ifx\csname urlstyle\endcsname\relax
  \providecommand{\doi}[1]{doi: #1}\else
  \providecommand{\doi}{doi: \begingroup \urlstyle{rm}\Url}\fi

\bibitem[{Black Forest Labs}(2025)]{FLUX2}
{Black Forest Labs}.
\newblock {FLUX.2}: Frontier visual intelligence.
\newblock \url{https://bfl.ai/blog/flux-2}, 2025.

\bibitem[Brateanu et~al.(2026)Brateanu, Mu, Ancuti, and Ancuti]{Multinex}
Alexandru Brateanu, Tingting Mu, Codruta~O. Ancuti, and Cosmin Ancuti.
\newblock Multinex: Lightweight low-light image enhancement via multi-prior
  retinex.
\newblock In \emph{Proceedings of the IEEE/CVF Conference on Computer Vision
  and Pattern Recognition}, pages 29887--29896, 2026.

\bibitem[Dong et~al.(2026)Dong, Li, Ren, Hu, Shi, Zhu, and Mou]{DATPRLIR}
Guanglu Dong, Chunlei Li, Chao Ren, Jingliang Hu, Yilei Shi, Xiaoxiang Zhu, and
  Lichao Mou.
\newblock Learning domain-aware task prompt representations for multi-domain
  all-in-one image restoration.
\newblock In \emph{Proceedings of the International Conference on Learning
  Representations}, pages 133754--133777, 2026.

\bibitem[Gabeur et~al.(2026)Gabeur, Long, Peng, Voigtlaender, Sun, Bao, Truong,
  Wang, Zhou, Barron, et~al.]{VisionBanana}
Valentin Gabeur, Shangbang Long, Songyou Peng, Paul Voigtlaender, Shuyang Sun,
  Yanan Bao, Karen Truong, Zhicheng Wang, Wenlei Zhou, Jonathan~T Barron,
  et~al.
\newblock Image generators are generalist vision learners.
\newblock \emph{arXiv preprint arXiv:2604.20329}, 2026.

\bibitem[Golestaneh et~al.(2022)Golestaneh, Dadsetan, and Kitani]{TReS}
S.~Alireza Golestaneh, Saba Dadsetan, and Kris~M. Kitani.
\newblock No-reference image quality assessment via transformers, relative
  ranking, and self-consistency.
\newblock In \emph{Proceedings of the IEEE/CVF Winter Conference on
  Applications of Computer Vision}, pages 3989--3999, 2022.

\bibitem[Guo et~al.(2022)Guo, Lu, Hou, Liu, Cheng, and Hu]{SegNeXt}
Meng-Hao Guo, Cheng-Ze Lu, Qibin Hou, Zhengning Liu, Ming-Ming Cheng, and
  Shi-Min Hu.
\newblock {SegNeXt}: Rethinking convolutional attention design for semantic
  segmentation.
\newblock \emph{Advances in Neural Information Processing Systems},
  35:\penalty0 1140--1156, 2022.

\bibitem[Ha et~al.(2017)Ha, Watanabe, Karasawa, Ushiku, and Harada]{MFNet}
Qishen Ha, Kohei Watanabe, Takumi Karasawa, Yoshitaka Ushiku, and Tatsuya
  Harada.
\newblock {MFNet}: Towards real-time semantic segmentation for autonomous
  vehicles with multi-spectral scenes.
\newblock In \emph{Proceedings of the IEEE/RSJ International Conference on
  Intelligent Robots and Systems}, pages 5108--5115, 2017.

\bibitem[Jia et~al.(2021)Jia, Zhu, Li, Tang, and Zhou]{LLVIP}
Xinyu Jia, Chuang Zhu, Minzhen Li, Wenqi Tang, and Wenli Zhou.
\newblock {LLVIP}: A visible-infrared paired dataset for low-light vision.
\newblock In \emph{Proceedings of the IEEE/CVF International Conference on
  Computer Vision Workshops}, pages 3489--3497, 2021.

\bibitem[Ke et~al.(2021)Ke, Wang, Wang, Milanfar, and Yang]{MUSIQ}
Junjie Ke, Qifei Wang, Yilin Wang, Peyman Milanfar, and Feng Yang.
\newblock {MUSIQ}: Multi-scale image quality transformer.
\newblock In \emph{Proceedings of the IEEE/CVF International Conference on
  Computer Vision}, pages 5128--5137, 2021.

\bibitem[Li et~al.(2024)Li, Li, Tu, Liu, Guo, Juefei-Xu, Xu, and Yu]{LightDiff}
Jinlong Li, Baolu Li, Zhengzhong Tu, Xinyu Liu, Qing Guo, Felix Juefei-Xu,
  Runsheng Xu, and Hongkai Yu.
\newblock Light the night: A multi-condition diffusion framework for unpaired
  low-light enhancement in autonomous driving.
\newblock In \emph{Proceedings of the IEEE/CVF Conference on Computer Vision
  and Pattern Recognition}, pages 15205--15215, 2024.

\bibitem[Liu et~al.(2023)Liu, Liu, Wu, Ma, Liu, Zhong, Luo, and Fan]{SegMiF}
Jinyuan Liu, Zhu Liu, Guanyao Wu, Long Ma, Risheng Liu, Wei Zhong, Zhongxuan
  Luo, and Xin Fan.
\newblock Multi-interactive feature learning and a full-time multi-modality
  benchmark for image fusion and segmentation.
\newblock In \emph{Proceedings of the IEEE/CVF International Conference on
  Computer Vision}, pages 8081--8090, 2023.

\bibitem[Ma et~al.(2022)Ma, Tang, Fan, Huang, Mei, and Ma]{SwinFusion}
Jiayi Ma, Linfeng Tang, Fan Fan, Jun Huang, Xiaoguang Mei, and Yong Ma.
\newblock {SwinFusion}: Cross-domain long-range learning for general image
  fusion via swin transformer.
\newblock \emph{IEEE/CAA Journal of Automatica Sinica}, 9\penalty0
  (7):\penalty0 1200--1217, 2022.

\bibitem[Tang et~al.(2025)Tang, Wang, Cai, Jiang, and Ma]{ControlFusion}
Linfeng Tang, Yeda Wang, Zhanchuan Cai, Junjun Jiang, and Jiayi Ma.
\newblock {ControlFusion}: A controllable image fusion network with
  language-vision degradation prompts.
\newblock \emph{Advances in Neural Information Processing Systems},
  38:\penalty0 118304--118327, 2025.

\bibitem[Tang et~al.(2026)Tang, Li, and Ma]{MaskDiFuser}
Linfeng Tang, Chunyu Li, and Jiayi Ma.
\newblock {Mask-DiFuser}: A masked diffusion model for unified unsupervised
  image fusion.
\newblock \emph{IEEE Transactions on Pattern Analysis and Machine
  Intelligence}, 48\penalty0 (1):\penalty0 591--608, 2026.

\bibitem[Wang et~al.(2023)Wang, Chan, and Loy]{CLIPIQA}
Jianyi Wang, Kelvin C.~K. Chan, and Chen~Change Loy.
\newblock Exploring clip for assessing the look and feel of images.
\newblock In \emph{Proceedings of the AAAI Conference on Artificial
  Intelligence}, pages 2555--2563, 2023.

\bibitem[Wang et~al.(2026{\natexlab{a}})Wang, Deng, Cao, and Vivone]{EFRUF}
Wu Wang, Liang-Jian Deng, Qi Cao, and Gemine Vivone.
\newblock A general image fusion approach exploiting gradient transfer learning
  and fusion rule unfolding.
\newblock \emph{IEEE Transactions on Pattern Analysis and Machine
  Intelligence}, 48\penalty0 (6):\penalty0 6347--6364, 2026{\natexlab{a}}.

\bibitem[Wang et~al.(2026{\natexlab{b}})Wang, Cui, Wang, Zhang, Wang, Zhang,
  Luo, Sun, Li, Wang, et~al.]{Emu3}
Xinlong Wang, Yufeng Cui, Jinsheng Wang, Fan Zhang, Yueze Wang, Xiaosong Zhang,
  Zhengxiong Luo, Quan Sun, Zhen Li, Yuqi Wang, et~al.
\newblock Multimodal learning with next-token prediction for large multimodal
  models.
\newblock \emph{Nature}, 650\penalty0 (8101):\penalty0 327--333,
  2026{\natexlab{b}}.

\bibitem[Xu et~al.(2022)Xu, Ma, Jiang, Guo, and Ling]{U2Fusion}
Han Xu, Jiayi Ma, Junjun Jiang, Xiaojie Guo, and Haibin Ling.
\newblock U2fusion: A unified unsupervised image fusion network.
\newblock \emph{IEEE Transactions on Pattern Analysis and Machine
  Intelligence}, 44\penalty0 (1):\penalty0 502--518, 2022.

\bibitem[Yang et~al.(2022)Yang, Wu, Shi, Lao, Gong, Cao, Wang, and
  Yang]{MANIQA}
Sidi Yang, Tianhe Wu, Shuwei Shi, Shanshan Lao, Yuan Gong, Mingdeng Cao, Jiahao
  Wang, and Yujiu Yang.
\newblock {MANIQA}: Multi-dimension attention network for no-reference image
  quality assessment.
\newblock In \emph{Proceedings of the IEEE/CVF Conference on Computer Vision
  and Pattern Recognition Workshops}, pages 1190--1199, 2022.

\bibitem[Yi et~al.(2024)Yi, Xu, Zhang, Tang, and Ma]{TextIF}
Xunpeng Yi, Han Xu, Hao Zhang, Linfeng Tang, and Jiayi Ma.
\newblock {Text-IF}: Leveraging semantic text guidance for degradation-aware
  and interactive image fusion.
\newblock In \emph{Proceedings of the IEEE/CVF Conference on Computer Vision
  and Pattern Recognition}, pages 27016--27025, 2024.

\bibitem[Yi et~al.(2026)Yi, Yan, Zhang, Li, Zhang, Cai, and Ma]{yi2026ll}
Xunpeng Yi, Qinglong Yan, Yibing Zhang, Zongrong Li, Hao Zhang, Zhanchuan Cai,
  and Jiayi Ma.
\newblock {LL-Refiner}: Learning adaptive refinement for ultra-high-definition
  low-light image enhancement.
\newblock \emph{IEEE/CAA Journal of Automatica Sinica}, 13\penalty0
  (6):\penalty0 1288--1300, 2026.

\bibitem[Zhang et~al.(2021)Zhang, Xu, Tian, Jiang, and Ma]{zhang2021image}
Hao Zhang, Han Xu, Xin Tian, Junjun Jiang, and Jiayi Ma.
\newblock Image fusion meets deep learning: A survey and perspective.
\newblock \emph{Information Fusion}, 76:\penalty0 323--336, 2021.

\bibitem[Zhang et~al.(2024)Zhang, Zuo, Jiang, Guo, and Ma]{MRFS}
Hao Zhang, Xuhui Zuo, Jie Jiang, Chunchao Guo, and Jiayi Ma.
\newblock {MRFS}: Mutually reinforcing image fusion and segmentation.
\newblock In \emph{Proceedings of the IEEE/CVF Conference on Computer Vision
  and Pattern Recognition}, pages 26964--26973, 2024.

\bibitem[Zhang et~al.(2025)Zhang, Cao, Zuo, Shao, and Ma]{OmniFuse}
Hao Zhang, Lei Cao, Xuhui Zuo, Zhenfeng Shao, and Jiayi Ma.
\newblock {OmniFuse}: Composite degradation-robust image fusion with
  language-driven semantics.
\newblock \emph{IEEE Transactions on Pattern Analysis and Machine
  Intelligence}, 47\penalty0 (9):\penalty0 7577--7595, 2025.

\bibitem[Zhang et~al.(2026{\natexlab{a}})Zhang, Gong, Wu, and
  Ma]{zhang2026dpafuse}
Hao Zhang, Meiqi Gong, Douyu Wu, and Jiayi Ma.
\newblock {DPAFuse}: A dual-space probabilistic adversarial image fusion
  framework with robust coding embedding.
\newblock \emph{Fundamental Research}, 6\penalty0 (4):\penalty0 2587--2601,
  2026{\natexlab{a}}.

\bibitem[Zhang et~al.(2026{\natexlab{b}})Zhang, Zha, Zhuang, Shao, and Ma]{RFC}
Hao Zhang, Yanping Zha, Qingwei Zhuang, Zhenfeng Shao, and Jiayi Ma.
\newblock Robust fusion controller: Degradation-aware image fusion with
  fine-grained language instructions.
\newblock In \emph{Proceedings of the AAAI Conference on Artificial
  Intelligence}, pages 12466--12474, 2026{\natexlab{b}}.

\bibitem[Zhang and Demiris(2023)]{zhang2023visible}
Xingchen Zhang and Yiannis Demiris.
\newblock Visible and infrared image fusion using deep learning.
\newblock \emph{IEEE Transactions on Pattern Analysis and Machine
  Intelligence}, 45\penalty0 (8):\penalty0 10535--10554, 2023.

\bibitem[Zhang et~al.(2026{\natexlab{c}})Zhang, Yi, Yan, Wang, Xu, and
  Ma]{UHDMFF}
Yibing Zhang, Xunpeng Yi, Qinglong Yan, Yeda Wang, Han Xu, and Jiayi Ma.
\newblock {UHD-MFF}: Shattering barriers in multi-focus ultra-high-definition
  image fusion via learnable lookup tables.
\newblock In \emph{Proceedings of the European Conference on Computer Vision},
  pages 294--311, 2026{\natexlab{c}}.

\bibitem[Zhao et~al.(2023)Zhao, Bai, Zhu, Zhang, Xu, Zhang, Zhang, Meng,
  Timofte, and Van~Gool]{DDFM}
Zixiang Zhao, Haowen Bai, Yuanzhi Zhu, Jiangshe Zhang, Shuang Xu, Yulun Zhang,
  Kai Zhang, Deyu Meng, Radu Timofte, and Luc Van~Gool.
\newblock {DDFM}: Denoising diffusion model for multi-modality image fusion.
\newblock In \emph{Proceedings of the IEEE/CVF International Conference on
  Computer Vision}, pages 8048--8059, 2023.

\end{thebibliography}
